\pdfoutput=1

\documentclass[11pt]{article}

\usepackage{acl}

\usepackage{times}
\usepackage{latexsym}

\usepackage[T1]{fontenc}

\usepackage[utf8]{inputenc}

\usepackage{microtype}

\usepackage{times}
\usepackage{latexsym}

\usepackage{microtype}

\usepackage{times}
\usepackage{latexsym}
\usepackage{booktabs}
\usepackage[T1]{fontenc}

\usepackage[utf8]{inputenc}

\usepackage{microtype}
\usepackage{amsmath}
\usepackage{booktabs}
\usepackage{array}
\newcolumntype{L}{>{$}l<{$}}
\newcolumntype{C}{>{$}c<{$}}
\newcolumntype{R}{>{$}r<{$}}

\usepackage{xcolor}

\usepackage{times}
\usepackage{latexsym}

\usepackage{microtype}

\usepackage{times}
\usepackage{latexsym}

\usepackage[T1]{fontenc}

\usepackage[utf8]{inputenc}

\usepackage{multirow}
\usepackage{array, tabularx, caption, boldline}
\usepackage{graphicx}
\usepackage{cellspace}
\usepackage{algpseudocode}  
\usepackage{amsmath}
\usepackage{multicol}  
\usepackage{multirow} 
\usepackage{flexisym}
\usepackage{graphicx}  
\usepackage{array}

\usepackage{microtype}

\usepackage{tabularx}
\makeatletter
\def\hlinewd#1{%
\noalign{\ifnum0=`}\fi\hrule \@height #1 %
\futurelet\reserved@a\@xhline}
\makeatother

\usepackage{times}
\usepackage{latexsym}

\usepackage{algpseudocode}  
\usepackage{amsmath}
\usepackage{multicol}  
\usepackage{multirow} 
\usepackage{graphicx}  
\usepackage{array}
\usepackage{arydshln}
\usepackage{booktabs}
\usepackage{textcomp}

\usepackage{amssymb}
\usepackage{pifont}

\usepackage{cleveref}
\crefformat{section}{\S#2#1#3} 
\crefformat{subsection}{\S#2#1#3}
\crefformat{subsubsection}{\S#2#1#3}

\usepackage{microtype}

\usepackage{adjustbox}
\usepackage{multicol}  
\usepackage{multirow} 
\usepackage{amsmath}
\usepackage{amsfonts}
\usepackage{makecell}
\usepackage{algpseudocode} 
\usepackage{verbatim}
\usepackage{mathtools}

\usepackage{booktabs}

\usepackage{makecell}
\usepackage{cleveref}
\usepackage[normalem]{ulem}

\usepackage{times}
\usepackage{latexsym}

\usepackage{footnote}
\makesavenoteenv{tabular}
\makesavenoteenv{table}

\usepackage[hang,flushmargin]{footmisc}

\usepackage[linesnumbered,ruled]{algorithm2e}
\SetAlFnt{\small}
\SetAlCapNameFnt{\normalsize}
\makeatletter
\newcommand{\nosemic}{\renewcommand{\@endalgocfline}{\relax}}
\newcommand{\dosemic}{\renewcommand{\@endalgocfline}{\algocf@endline}}
\let\oldnl\nl
\newcommand{\nonl}{\renewcommand{\nl}{\let\nl\oldnl}}
\makeatother

\usepackage{enumitem}
\usepackage{tablefootnote}

\newcommand*\samethanks[1][\value{footnote}]{\footnotemark[#1]}

\usepackage{tabularx}
\makeatletter
\def\hlinewd#1{%
\noalign{\ifnum0=`}\fi\hrule \@height #1 %
\futurelet\reserved@a\@xhline}
\makeatother

\title{Multi-Task Pre-Training for Plug-and-Play\\ Task-Oriented Dialogue System}

\author{Yixuan Su$^{\spadesuit,}$\thanks{~~Work done during authors’ internship at Amazon.}~\quad Lei Shu$^{\heartsuit}$\quad Elman Mansimov$^{\heartsuit}$\quad Arshit Gupta$^{\heartsuit}$\quad \\\textbf{Deng Cai}$^{\clubsuit,}$\samethanks \quad  \textbf{Yi-An Lai}$^{\heartsuit}$\quad  \textbf{Yi Zhang}$^{\heartsuit}$\\

$^{\spadesuit}$University of Cambridge \ \ \ \ \ $^\heartsuit$Amazon AWS AI \\
$^\clubsuit$The Chinese University of Hong Kong\\
{\tt ys484@cam.ac.uk, thisisjcycd@gmail.com}\\ 
{\tt \{leishu,mansimov,arshig,yianl,yizhngn\}@amazon.com}\\
}

\begin{document}
\maketitle
\begin{abstract}
Pre-trained language models have been recently shown to benefit task-oriented dialogue (TOD) systems. Despite their success, existing methods often formulate this task as a cascaded generation problem which can lead to error accumulation across different sub-tasks and greater data annotation overhead. In this study, we present PPTOD, a unified plug-and-play model for task-oriented dialogue. In addition, we introduce a new dialogue multi-task pre-training strategy that allows the model to learn the primary TOD task completion skills from heterogeneous dialog corpora. We extensively test our model on three benchmark TOD tasks, including end-to-end dialogue modelling, dialogue state tracking, and intent classification. Experimental results show that PPTOD achieves new state of the art on all evaluated tasks in both high-resource and low-resource scenarios. Furthermore, comparisons against previous SOTA methods show that the responses generated by PPTOD are more factually correct and semantically coherent as judged by human annotators.\footnote{Our code, models and other related resources are publicly available at \url{https://github.com/awslabs/pptod}}
\end{abstract}

\section{Introduction}
Task-oriented dialogue is often decomposed into three sub-tasks: (1) dialogue state tracking (DST) for tracking user's belief state; (2) dialogue policy learning (POL) for deciding which system action to take; (3) natural language generation (NLG) for generating dialogue response \cite{DBLP:journals/pieee/YoungGTW13}. 

Traditional approaches \citep{10.5555/203290,DBLP:journals/pieee/YoungGTW13} adopt a modularized pipeline that addresses different sub-tasks with distinct dedicated modules. In contrast, recent systems \citep{wen-etal-2017-network,eric-etal-2017-key,lei-etal-2018-sequicity, shu2019flexibly} integrate all functionalities required to hold a dialogue into neural network models. With the advances in pre-trained language models (PLMs) \citep{radford2019language,DBLP:conf/naacl/DevlinCLT19,DBLP:journals/jmlr/RaffelSRLNMZLL20}, different systems based on PLMs have been proposed \citep{DBLP:conf/nips/Hosseini-AslMWY20,DBLP:conf/emnlp/LinMWF20,peng2021soloist,DBLP:journals/tacl/LiuYRB21}. 
Despite their differences, most existing methods formulate task-oriented dialogue as a cascaded generation problem, that is, the model can only solve latter sub-tasks by conditioning on the outputs of previous ones. For instance, to generate the response (NLG), the model must rely on the outputs of previous sub-tasks (i.e., DST and POL).

While impressive results are reported \cite{DBLP:conf/nips/Hosseini-AslMWY20,peng2021soloist},  
we identify three major limitations in the cascaded formulation of their system design. (1) Firstly, as the model solves all sub-tasks in a sequential order, the errors accumulated from previous steps are propagated to latter steps \citep{DBLP:journals/corr/LiCLGC17,DBLP:conf/naacl/LiuL18}. (2) Secondly, the training data must be annotated for all sub-tasks. Such annotation requirement significantly increases the data curation overhead. More importantly, it precludes the model from using the large amount of existing data that is partially annotated (e.g., data only annotated with DST or NLG). (3) Thirdly, the results of different sub-tasks must be generated in a cascaded order which inevitably increases the system inference latency.

\begin{figure*}[h] 
  \centering    
  \setlength{\abovecaptionskip}{3pt}
  \includegraphics[width=1.0\textwidth]{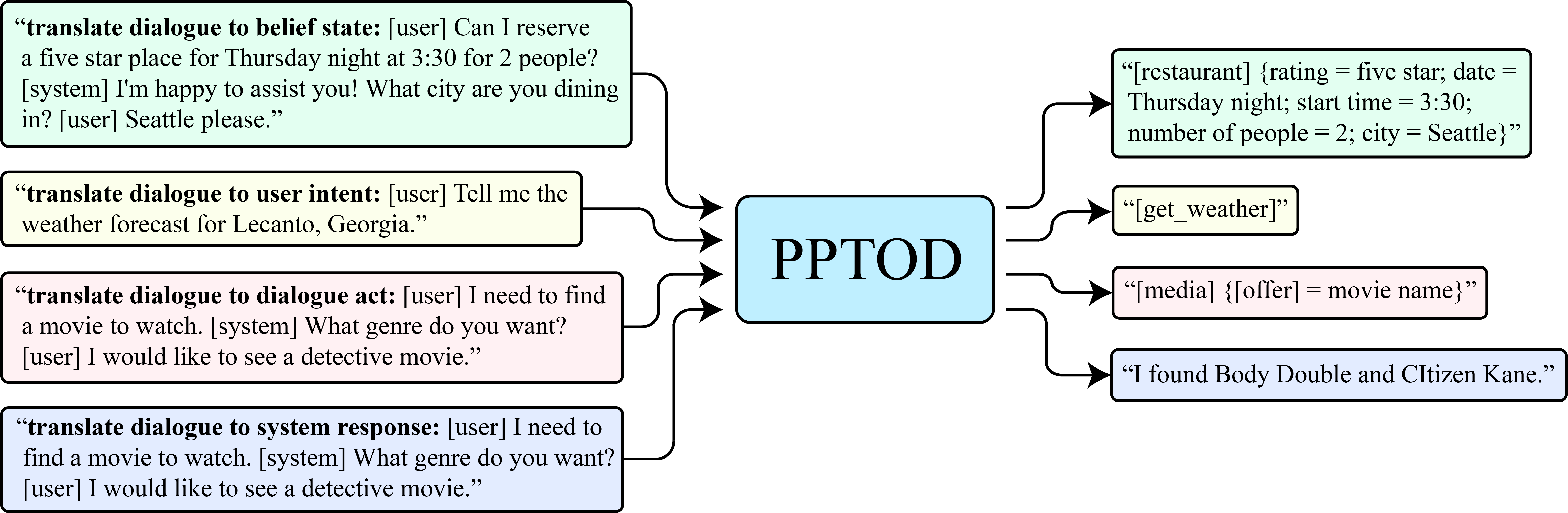}
  \caption{\textbf{Overview}: In the dialogue multi-task pre-training stage, we pre-train our model with four TOD-related tasks, including natural language understanding (NLU), dialogue state tracking (DST), dialogue policy learning (POL), and natural language generation (NLG). For each task, the model takes the dialogue context and the task-specific prompt as input and learns to generate the corresponding target text. Our learning framework allows us to train the model with partially annotated data across a diverse set of tasks. 
  (best viewed in color)}
  \label{fig:overview}
\end{figure*}

In this study, we propose a novel  \textbf{P}lug-and-\textbf{P}lay \textbf{T}ask-\textbf{O}riented \textbf{D}ialogue (PPTOD) system. 
Figure \ref{fig:overview} depicts an illustration of our approach. As seen, we integrate different dialogue modules (e.g. DST, POL, and NLG) into a unified model. Motivated by the concept of \textit{in-context learning} 
\citep{DBLP:conf/nips/BrownMRSKDNSSAA20}, to steer the model to solve different TOD sub-task, we plug a task-specific natural language instruction, termed as \textit{prompt}, into the dialogue context as the model input. This way, the generations of different sub-tasks are decoupled, leading to a greater flexibility of the model that brings us at least two advantages: (1) As different sub-tasks are solved separately, the model can learn from data that is partially annotated for different sub-tasks (e.g., DST and NLG). (2) The outputs of different sub-tasks are generated in parallel which alleviates the problem of error accumulation and reduces the system inference latency. 

Inspired by recent success of dialogue language model pre-training \cite{DBLP:conf/acl/ZhangSGCBGGLD20,wu-etal-2020-tod,peng2021soloist}, we propose a dialogue multi-task pre-training strategy that equips our model with the primary TOD task completion skills. Specifically, initialized with T5 \cite{DBLP:journals/jmlr/RaffelSRLNMZLL20}, we pre-train our model on a heterogeneous set of dialog corpora that consist of partially-annotated data. To build the pre-training corpora, we collect and combine eleven human-written multi-turn dialogue corpora. The collected datasets are partially annotated for some of the TOD-related tasks, including natural language understanding (NLU), dialogue state tracking (DST), dialogue policy learning (POL), and natural language generation (NLG). 
In total, the pre-training corpora contain over 2.3M utterances across over 80 domains (see more details in Table \ref{tb:pretraining_corpus}). When applying the pre-trained PPTOD to a new task, we fine-tune it using the same learning objective as in the pre-training stage.

We evaluate PPTOD on a wide range of benchmark TOD tasks, including end-to-end dialogue modelling, dialogue state tracking, and intent classification.  Comparisons against previous state-of-the-art approaches show that PPTOD achieves better performance in both full-training and low-resource settings as judged by automatic and human evaluations. In summary, our contributions are: 


\begin{itemize}
    \item A novel model, PPTOD, that effectively leverages pre-trained language models for task-oriented dialogue tasks.
    \item A new dialogue multi-task pre-training strategy that augments the model's ability with heterogeneous dialogue corpora.
    \item  Extensive evaluations on three benchmark TOD tasks reporting state-of-the-art results in both full-training and low-resource settings.    
    \item In-depth analysis that further reveals the merits of our model design and the proposed multi-task pre-training strategy.
\end{itemize}

\section{Related Work}
\paragraph{Task-Oriented Dialogue.} Task-oriented dialogue aims at accomplishing  user's goal. 
Traditional systems \cite{DBLP:journals/csl/WilliamsY07,DBLP:journals/pieee/YoungGTW13} adopt a pipelined approach that requires dialogue state tracking for understanding user's goal, dialogue policy learning for deciding which system action to take, and natural language generation for generating dialogue responses. 

Recently, to simplify the modelling effort, researchers have shifted their attention to building neural network models that address the TOD sub-tasks \cite{wen-etal-2017-network,eric-etal-2017-key,lei-etal-2018-sequicity,DBLP:conf/aaai/LiangTCY20}. 
With the advances in pre-trained language models (PLMs), \citet{budzianowski-vulic-2019-hello} first applied the GPT-2 model for the NLG task. \citet{DBLP:conf/emnlp/LinMWF20} and \citet{DBLP:conf/aaai/YangLQ21} moved one step forward and utilized pre-trained language models to solve all TOD sub-tasks conditioned on the history of oracle belief states. Based on the GPT-2 model, \citet{DBLP:conf/nips/Hosseini-AslMWY20} proposed a cascaded model, SimpleTOD, that addresses all TOD sub-tasks without using the oracle information. To improve the system performance, \citet{peng2021soloist} and \citet{DBLP:journals/tacl/LiuYRB21} applied dialogue pre-training over external dialogue corpora. However, both methods require the pre-training data to be fully annotated for all TOD sub-tasks (i.e., DST, POL, and NLG) which greatly limits the amount of data they can use. Additionally,  \citet{DBLP:journals/tacl/LiuYRB21} achieved better results with noisy chanel model that requires two additional language models for outputs re-scoring. Unlike their approach, we address the task of task-oriented dialogue with a single unified model. Lastly, concurrent work by \citet{DBLP:journals/corr/abs-2111-14592} shows that adding an unified dialogue act prediction task for policy optimization helps to improve the performance of the pre-trained task-oriented dialogue model.


\paragraph{Language Model Pre-training.} The research community has witnessed remarkable progress of pre-training methods in a wide range
of NLP tasks, including language understanding
\cite{DBLP:conf/naacl/PetersNIGCLZ18,DBLP:conf/naacl/DevlinCLT19,DBLP:journals/corr/abs-1907-11692,DBLP:conf/nips/YangDYCSL19,DBLP:journals/corr/abs-2111-04198} and text generation \cite{radford2019language,DBLP:conf/acl/LewisLGGMLSZ20,DBLP:journals/jmlr/RaffelSRLNMZLL20,DBLP:journals/taslp/SuWCBKC21,DBLP:conf/emnlp/SuVWFC21,DBLP:conf/emnlp/SuMBC21,DBLP:journals/corr/abs-2202-06417}.

In the dialogue domain, many models are pre-trained on open-domain conversational data like Reddit. Based on GPT-2, Transfertransfo \cite{DBLP:journals/corr/abs-1901-08149} achieves good results on ConvAI-2 competition. As another extension of GPT-2, DialoGPT \cite{DBLP:conf/acl/ZhangSGCBGGLD20} performs well in generating open-domain dialogue response. ConveRT \cite{DBLP:conf/emnlp/HendersonCMSWV20} is a language model with dual-encoder built for the task of response selection.  PLATO \cite{DBLP:conf/acl/BaoHWWW20} pre-trains a model with discrete latent variable structure for the response generation task.  \citet{wu-etal-2020-tod} adapts BERT with TOD pre-training and achieves strong performances on four dialogue understanding tasks. 

\paragraph{Pre-training on Supplementary Data.} Recent work \citep{phang2018stilts,DBLP:armen2021muppet} found that supplementary training on the tasks with intermediate-labelled data improves the performance of the fine-tuned models on GLUE natural language understanding benchmark~\citep{DBLP:wang2019glue}. Our work studies a similar supplementary training setup with intermediate-labelled data for task-oriented dialogue systems. Unlike previous work, we use a single multi-task model for all relevant sub-tasks in task-oriented dialogue systems.


\section{Methodology}
In this section, we first discuss the datasets and learning objective used in the proposed dialogue multi-task pre-training. Then we introduce how to apply the pre-trained PPTOD for a new task.

\begin{table}[t]
    \small
	\centering  
	\renewcommand{\arraystretch}{1.2}
	\setlength{\tabcolsep}{6pt}
	\scalebox{0.83}{
	\begin{tabular}{ccccccc}
		\hlinewd{0.75pt}
		\multirow{2}{*}{\textbf{Dataset}}&\multicolumn{4}{c}{\textbf{Data Annotation}}&\multirow{2}{*}{\textbf{Utter.}}&\multirow{2}{*}{\textbf{Dom.}}\\
		\cmidrule(lr){2-5}
		&NLU&DST&POL&NLG&&\\
		\hline
		MetaLWOZ&$\times$&$\times$&$\times$&$\checkmark$&822,932&47\\
		SNIPS&$\checkmark$&$\times$&$\times$&$\times$&25,682&9\\
		CLINC&$\checkmark$&$\times$&$\times$&$\times$&45,000&10\\
		ATIS&$\checkmark$&$\times$&$\times$&$\times$&10,772&1\\
        KVRET&$\times$&$\checkmark$&$\times$&$\checkmark$&31,504&3\\
        WOZ&$\times$&$\checkmark$&$\times$&$\checkmark$&15,248&1\\
        CamRest676&$\times$&$\checkmark$&$\times$&$\checkmark$&10,976&1\\
        MSR-E2E&$\times$&$\checkmark$&$\checkmark$&$\checkmark$&72,238&3\\
        Frames&$\times$&$\checkmark$&$\checkmark$&$\checkmark$&38,316&1\\
        TaskMaster&$\times$&$\checkmark$&$\checkmark$&$\checkmark$&540,688&6\\
        Schema-Guided&$\times$&$\checkmark$&$\checkmark$&$\checkmark$&757,380&17\\
		\hlinewd{0.75pt}
	\end{tabular}}
    \caption{The summary of data annotations and number of utterances (Utter.) as well as domains (Dom.) for all pre-training corpora. All datasets are partially annotated for some of the TOD-related tasks, including natural language understanding (NLU), dialogue state tracking (DST), dialogue policy learning (POL), and natural language generation (NLG).}
	\label{tb:pretraining_corpus}
\end{table}

\subsection{Pre-training Datasets}
To construct the pre-training corpus, we collect eleven human-written multi-turn task-oriented dialogue corpora, including MetaLWOZ \cite{lee2019multi-domain}, SNIPS \cite{DBLP:journals/corr/abs-1805-10190}, CLINC \cite{larson-etal-2019-evaluation}, ATIS \cite{ATIS}, KVRET \cite{eric-etal-2017-key}, WOZ \cite{mrksic-etal-2017-neural}, MSR-E2E \cite{li2018microsoft}, Frames \cite{el-asri-etal-2017-frames}, TaskMaster \cite{DBLP:conf/emnlp/ByrneKSNGDYDKC19}, and Schema-Guided \cite{DBLP:conf/aaai/RastogiZSGK20}. In total, there are over 2.3M utterances across 80 domains. In Table \ref{tb:pretraining_corpus}, we provide the details of data annotations and utterance/domain statistics of all datasets.\footnote{More dataset descriptions are provided in Appendix \ref{dataset_details}.}

\subsection{Dialogue Multi-Task Pre-training}
\label{sec:multi-task-pretraining}
Motivated by previous work \cite{DBLP:journals/corr/abs-1806-08730,DBLP:journals/corr/abs-1904-09286,DBLP:journals/jmlr/RaffelSRLNMZLL20} that unify multiple NLP tasks into a common format, we cast all TOD-related tasks that we consider into the same plug-and-play text generation problem. To specify the target task, we plug a task-specific prompt into the dialogue context as the model input. Figure \ref{fig:overview} depicts an illustration of our approach.

In the multi-task pre-training stage, each training sample is represented as:
\begin{equation}
    d = (z_t, x, y), 
\end{equation} 
where $t$ denotes the TOD task that the sample $d$ belongs to, and $t\in\{\textup{NLU},\textup{DST},\textup{POL},\textup{NLG}\}$. 
$z_t$ is the task-specific prompt of the form ``\texttt{translate dialogue to A:}'', with \texttt{A} corresponding to ``user intent'', ``belief state'', ``dialogue act'', and ``system response'' for the tasks of NLU, DST, POL, and NLG, respectively. 
$x$ denotes the input dialogue context which is a concatenation of all previous utterances in the dialogue - both system's and user's. 
And $y$ denotes the target output text. 

As an example presented in Figure \ref{fig:overview}, to perform the user intent classification task (i.e., NLU), the model is fed with the sequence \textit{``translate dialogue to user intent: [user] Tell me the weather forecast for Lecanto, Georgia.''} and is trained to generate the user intent label text \textit{``[get\_weather]''}.

\normalem
\begin{algorithm}[t]
   \SetAlCapFnt{\small}
   \SetAlCapNameFnt{\small}
    \caption{Dialogue Multi-Task Pre-Training}
    \SetKwInOut{Input}{Input}
    \SetKwInOut{Output}{Output}
    
    \Input{Dataset $\mathcal{D}=\{(z_t, x, y)_i\}_{i=1}^{|\mathcal{D}|}$; model trainer $\mathcal{T}$ that takes batches of training data as input to optimize the model parameters $\Theta$; maximum number of epochs $e_{\textup{max}}$;}
    \For{epoch $e = 1, ..., e_{\textup{max}}$}
    {
        Shuffle $\mathcal{D}$\ by mixing data from different tasks;
        \For{$B$ in $\mathcal{D}$}
          {
            Invoke trainer $\mathcal{T}$, using one batch of training data $B=\{(z_t, x, y)_k\}_{k=1}^{|B|}$ as input to optimize the model using $\mathcal{L}_{\Theta}$ (Eq. \eqref{eq:mle}).
          }
    }
    \Output{Trained Model $\Theta$}
\end{algorithm}

\paragraph{Learning.} The model is trained with a maximum likelihood 
objective. Given the training sample $d = (z_t, x, y)$, the objective $\mathcal{L}_{\Theta}$ is defined as
\begin{equation}
    \label{eq:mle}
    \mathcal{L}_{\Theta} = -\sum_{i=1}^{|y|}\log P_{\Theta}(y_i|y_{<i}; z_t, x),
\end{equation}
where $\Theta$ is the model parameters.  

In the multi-task pre-training stage, the model is trained to perform all TOD-related tasks with data annotated for different tasks. To optimize the model parameters $\Theta$, we use mini-batch based optimization approach as shown in Algorithm 1. 

\subsection{Fine-Tuning to a New Task}
When applying the pre-trained PPTOD to a new downstream task with task-specific labelled data, we use the same learning objective Eq. \eqref{eq:mle} as in the dialogue multi-task pre-training stage.

\begin{table*}[tb]
    \small
	\centering  
	\renewcommand{\arraystretch}{1.2}
	\setlength{\tabcolsep}{6pt}
	\scalebox{0.85}{
	\begin{tabular}{ccccccccc}
		\hlinewd{0.75pt}
		\multirow{2}{*}{\textbf{Model}}&\multicolumn{4}{c}{MultiWOZ 2.0}&\multicolumn{4}{c}{MultiWOZ 2.1}\\
		\cmidrule(lr){2-5}
		\cmidrule(lr){6-9}
		&Inform&Success&BLEU&Combined Score&Inform&Success &BLEU&Combined Score\\
		\hline
		Sequicity&66.41&45.32&15.54&71.41&-&-&-&-\\
		MD-Sequicity&75.72&58.32&15.40&82.40&-&-&-&-\\
		DAMD&76.33&60.40&16.60&84.97&&&&\\
		MinTL$\dagger$ &84.88&74.91&17.89&97.78&-&-&-&-\\  
        HIER-Joint &80.50&71.70&19.74&95.84&-&-&-&-\\
        SOLOIST&85.50&72.90&16.54&95.74&-&-&-&-\\
        TOP$\mathsection$&85.20&72.90&17.00&96.05&-&-&-&-\\ 
        TOP+NOD$\mathsection$&86.90&76.20&\textbf{20.58}&102.13&-&-&-&-\\
        LABES-S2S&-&-&-&-&78.07&67.06&18.13&90.69\\
		UBAR${\dagger,\ddagger}$&85.10&71.02&16.21&94.27&86.20&70.32&16.48&94.74\\
		SimpleTOD&84.40&70.10&15.01&92.26&85.00&70.50&15.23&92.98\\
		\hline
		$\textup{PPTOD}_{\textup{small}}$&87.80&75.30&19.89&101.44&\textbf{88.89}&76.98&18.59&101.52\\
		$\textup{PPTOD}_{\textup{base}}$&\textbf{89.20}&\textbf{79.40}&18.62&\textbf{102.92}&87.09&\textbf{79.08}&\textbf{19.17}&\textbf{102.26}\\
		$\textup{PPTOD}_{\textup{large}}$&82.60&74.10&19.21&97.56&86.43&74.35&17.89&98.28\\
		\hlinewd{0.75pt}
	\end{tabular}}
    \caption{End-to-end evaluation. $\dagger$: the models require the history of oracle dialogue states when making predictions at current turn. $\ddagger$: UBAR scores are acquired with the author-released models. $\mathsection$: as the authors did not release their code, we cite the results of TOP and TOP+NOD on MultiWOZ 2.0 from the original paper \cite{DBLP:journals/tacl/LiuYRB21}. 
    }
	\label{tb:full_training_multi_woz}
\end{table*}

\begin{table*}[tb]
    \small
	\centering  
	\renewcommand{\arraystretch}{1.2}
	\setlength{\tabcolsep}{6pt}
	\scalebox{0.72}{
	\begin{tabular}{ccccccccccccccccc}
		\hlinewd{0.75pt}
		\multirow{2}{*}{\textbf{Model}}&\multicolumn{4}{c}{1\% of training data}&\multicolumn{4}{c}{5\% of training data}&\multicolumn{4}{c}{10\% of training data}&\multicolumn{4}{c}{20\% of training data}\\
		\cmidrule(lr){2-5}
		\cmidrule(lr){6-9}
		\cmidrule(lr){10-13}
		\cmidrule(lr){14-17}
		&Inform&Succ.&BLEU&Comb.&Inform&Succ.&BLEU&Comb.&Inform&Succ.&BLEU&Comb.&Inform&Succ.&BLEU&Comb.\\
		\hline
		MD-Sequicity$\ddagger$&-&-&-&-&49.40&19.70&10.30&44.85&58.10&34.70&11.40&57.80&64.40&42.10&13.00&66.25\\
		DAMD$\dagger$&34.40&9.10&8.10&29.85&52.50&31.80&11.60&53.75&55.30&30.30&13.00&55.80&62.60&44.10&14.90&68.25\\
		SOLOIST$\dagger$&58.40&35.30&10.58&57.43&69.30&52.30&11.80&72.60&69.90&51.90&14.60&75.50&74.00&60.10&15.25&82.29\\  
		MinTL$\ddagger$&-&-&-&-&75.48&60.96&13.98&82.20&78.08&66.87&15.46&87.94&82.48&68.57&13.00&88.53\\
		\hline
		$\textup{PPTOD}_{\textup{small}}$&66.96&50.90&12.51&71.44&76.58&61.60&\textbf{15.35}&84.44&83.50&68.18&15.56&91.01&82.96&69.90&\textbf{17.02}&93.45\\
		$\textup{PPTOD}_{\textup{base}}$&\textbf{74.42}&\textbf{52.44}&\textbf{12.99}&\textbf{76.41}&\textbf{79.86}&\textbf{63.48}&14.89&\textbf{86.55}&\textbf{84.42}&\textbf{68.36}&\textbf{15.57}&\textbf{91.96}&\textbf{84.94}&71.70&17.01&\textbf{95.32}\\
		$\textup{PPTOD}_{\textup{large}}$&64.38&51.94&11.84&70.01&75.20&61.94&14.17&82.54&80.64&66.74&15.25&88.94&81.74&\textbf{72.18}&15.13&92.09\\
		\hlinewd{0.75pt}
	\end{tabular}}
    \caption{Low-resource evaluation on MultiWOZ 2.0, where Succ. and Comb. denote the Success and Combined Score metrics, respectively. $\ddagger$ and $\dagger$ results are cited from \citet{DBLP:conf/emnlp/LinMWF20} and \citet{peng2021soloist}.}
	\label{tb:few_shot_multi_woz_2_0_experiment}
\end{table*}

\subsection{Implementation Details}
In this work, we report results of PPTOD with three model sizes: $\textup{PPTOD}_{\textup{small}}$, $\textup{PPTOD}_{\textup{base}}$, and $\textup{PPTOD}_{\textup{large}}$. These three models are initialized with T5-small, T5-base, and T5-large models \cite{DBLP:journals/jmlr/RaffelSRLNMZLL20} that contain $\sim$60M, $\sim$220M, and $\sim$770M parameters, respectively. We pre-train the model with different configurations on our collected pre-training corpora for 10 epochs. The training samples are truncated to ensure a maximal length of 1024. 
The models are trained using Adam optimizer \cite{DBLP:journals/corr/KingmaB14} with a learning rate of 5e-5 and a batch size of 128. 
Our implementation is based on the Huggingface Library \cite{DBLP:journals/corr/abs-1910-03771}.

\section{Experiments}
We test PPTOD on three benchmark TOD tasks: (1) end-to-end dialogue modelling; (2) dialogue state tracking; and (3) user intent classification. 

\subsection{End-to-End Dialogue Modelling}
End-to-end dialogue modelling aims at evaluating the model in the most realistic, fully end-to-end setting, where the generated dialogue states are used for the database search and response generation \cite{DBLP:conf/aaai/ZhangOY20,DBLP:conf/nips/Hosseini-AslMWY20}.


\subsubsection{Dataset and Evaluation Metric}
We conduct experiments on the benchmark MultiWOZ 2.0 \cite{DBLP:conf/emnlp/BudzianowskiWTC18} and 2.1 \cite{DBLP:conf/lrec/EricGPSAGKGKH20} datasets.\footnote{Note that, there is no overlap between the MultiWOZ dataset and our dialogue pre-training corpora.} In MultiWOZ, the generation of response is not only related to the dialogue context,  
but also grounded on the database (DB) state. The DB state is automatically retrieved from a pre-defined database using the generated dialogue state (DST). Following previous studies, during inference, PPTOD first predicts the DST result to retrieve the DB state. Then, based on the retrieved DB state and the dialogue context, the results of POL and NLG are generated in parallel. In Section \cref{sec:analysis}, we further compare the performance of our model with or without using the DB state as input. 

For evaluation, we follow the original MultiWOZ guidance for all individual metrics: \textbf{Inform}, \textbf{Success}, and \textbf{BLEU} \cite{papineni-etal-2002-bleu}. An overall measurement, i.e., combined score \cite{DBLP:conf/sigdial/MehriSE19}, is also reported which is defined as \textbf{Combined} = (Inform + Success) $\times$ 0.5 + BLEU.

\subsubsection{Baselines}
We compare PPTOD with several strong baselines, including Sequicity \cite{lei-etal-2018-sequicity}, MD-Sequicity \cite{DBLP:conf/aaai/ZhangOY20}, DAMD \cite{DBLP:conf/aaai/ZhangOY20}, MinTL \cite{DBLP:conf/emnlp/LinMWF20}, HIER-Joint \cite{DBLP:conf/naacl/SantraAG21}, LABES-S2S \cite{DBLP:conf/emnlp/ZhangOHF20}, SimpleTOD \cite{DBLP:conf/nips/Hosseini-AslMWY20}, UBAR \cite{DBLP:conf/aaai/YangLQ21}, and SOLOIST \cite{peng2021soloist}, TOP and TOP+Noisy Online Decoding (TOP+NOD) \cite{DBLP:journals/tacl/LiuYRB21}.

\subsubsection{Full Training Evaluation}
Table \ref{tb:full_training_multi_woz} shows the main results. On both MultiWOZ 2.0 and 2.1 datasets, PPTOD performs better than previous SOTA methods on seven out of eight metrics. In particular, it is worth mentioning that our model is a single architecture that does not require additional language models for re-ranking the outputs as in TOP+NOD~\cite{DBLP:journals/tacl/LiuYRB21}. 
Moreover, the results show that the large size $\textup{PPTOD}_{\textup{large}}$ underperforms $\textup{PPTOD}_{\textup{small}}$ and $\textup{PPTOD}_{\textup{base}}$. Our analysis is that the large size model is less capable when learning to generate the delexicalized tokens, which are not seen during its pre-training stage, for the NLG task.



\subsubsection{Low-Resource Evaluation}
To investigate the generalization ability of PPTOD, we evaluate it in a more challenging low-resource scenario. Following previous studies, we train our model on MultiWOZ 2.0 by varying the percentage of training data, ranging from 1\% ($\sim$80 samples) to 20\% ($\sim$1600 samples). We compare our model with several strong baselines, including MD-Sequicity, DAMD, SOLOIST, and MinTL.\footnote{We did not compare results with TOP+NOD \cite{DBLP:journals/tacl/LiuYRB21} since the authors did not release their code and models.} 

In each low-resource setting, we train our model five times with different random seeds and different selection of training data. The average scores over five runs are presented in Table \ref{tb:few_shot_multi_woz_2_0_experiment}.\footnote{Detailed numerical results can be found in Appendix \ref{sec:few-shot_mwoz_2_0_complete_results}.} As seen, PPTOD consistently outperforms all baseline models by a large margin. Notably, our performance gain is even larger when fewer samples are used for training. This indicates that PPTOD better leverages the prior knowledge from pre-training therefore achieving better results in the extreme low-resource settings. Furthermore, with 20\% of training data, PPTOD can achieve results that are comparable to the scores of systems like SOLOIST that are trained with full dataset as reported in Table \ref{tb:full_training_multi_woz}. 

\begin{table}[tb]
    \small
	\centering  
	\renewcommand{\arraystretch}{1.2}
	\setlength{\tabcolsep}{6pt}
	\scalebox{0.82}{
	\begin{tabular}{ccc}
		\hlinewd{0.75pt}
		\multirow{2}{*}{\textbf{Model}}&\multicolumn{2}{c}{MWOZ Joint Acc.(\%)}\\
		\cmidrule(lr){2-3}
		&2.0&2.1\\
		\hline
		\multicolumn{3}{c}{\textit{{Classification-based Approaches}}}\\
		\hline
		GLAD \cite{DBLP:conf/acl/SocherZX18}&35.57&-\\
		GCE \cite{DBLP:journals/corr/abs-1812-00899}&36.27&-\\
		FJST \cite{DBLP:conf/lrec/EricGPSAGKGKH20}&40.20&38.00\\
		SUMBT \cite{DBLP:conf/acl/LeeLK19}&46.65&-\\
	   	TOD-BERT \cite{wu-etal-2020-tod}&-&48.00\\
		DS-Picklist \cite{DBLP:journals/corr/abs-1910-03544} ${\dagger}$&54.39&53.30\\
		SST \cite{DBLP:conf/aaai/0002LWZT020} ${\dagger}$&51.17&55.23\\
		TripPy \cite{DBLP:conf/sigdial/HeckNLGLMG20}&-&55.29\\
		CHAN \cite{DBLP:conf/acl/ShanLZMFNZ20} ${\dagger}$&52.68&58.55\\
		FPDSC-turn \cite{DBLP:journals/corr/abs-2107-05168} ${\dagger}$&\textbf{55.03}& 57.88\\
		FPDSC-dual \cite{DBLP:journals/corr/abs-2107-05168} ${\dagger}$&53.17&\textbf{59.07}\\
		
		\hline
		\multicolumn{3}{c}{\textit{{Generation-based Approaches}}}\\
		\hline
		Neural Reading \cite{DBLP:conf/sigdial/GaoSACH19}&41.10&-\\
		TRADE \cite{DBLP:conf/acl/WuMHXSF19}&48.62&46.00\\
		COMER \cite{DBLP:conf/emnlp/RenNM19}&48.79&-\\
	   	DSTQA \cite{DBLP:journals/corr/abs-1911-06192} ${\dagger}$&51.44&51.17\\
		SOM-DST \cite{DBLP:conf/acl/KimYKL20}&51.38&52.57\\
		LABES-S2S \cite{DBLP:conf/emnlp/ZhangOHF20}&-&51.45\\
		MinTL \cite{DBLP:conf/emnlp/LinMWF20}&52.10&53.62\\
		SimpleTOD \cite{DBLP:conf/nips/Hosseini-AslMWY20}&-&55.76\\
		Seq2seq-DU \cite{DBLP:conf/acl/FengWL20}&-&56.10\\
		UBAR \cite{DBLP:conf/aaai/YangLQ21}&52.59&56.20\\
		SOLOIST \cite{peng2021soloist}&53.20&56.85\\
		\hline
		$\textup{PPTOD}_{\textup{small}}$&51.50&56.47\\
		$\textup{PPTOD}_{\textup{base}}$&53.37&57.10\\
		$\textup{PPTOD}_{\textup{large}}$&\textbf{53.89}&\textbf{57.45}\\
		\hlinewd{0.75pt}
	\end{tabular}}
    \caption{DST results. ${\dagger}$: the models require a full pre-defined ontology for all possible domain-slot pairs.}
	\label{tb:dialogue_state_tracking}
\end{table}

\subsection{Dialogue State Tracking}
Next, we evaluate PPTOD for the dialogue state tracking task. The experiments are conducted on the benchmark MultiWOZ 2.0 \cite{DBLP:conf/emnlp/BudzianowskiWTC18} and 2.1 \cite{DBLP:conf/lrec/EricGPSAGKGKH20} datasets. For evaluation, the joint goal accuracy is reported.


\subsubsection{Full Training Evaluation}
We compare PPTOD with a wide range of existing methods that can be categorized into two classes: (1) classification-based approaches and (2) generation-based approaches. Table \ref{tb:dialogue_state_tracking} shows the DST results. Compared to other generation-based approaches, $\textup{PPTOD}_{\textup{large}}$ obtains the highest accuracy on both datasets. The performance of our model is lower than the SOTA classification-based approaches. However, these methods operate on a fixed ontology and perform prediction over a pre-defined set of slot-value pairs \cite{DBLP:journals/corr/abs-1910-03544,DBLP:conf/aaai/0002LWZT020,DBLP:conf/acl/ShanLZMFNZ20,DBLP:journals/corr/abs-2107-05168}. This idea of fixed ontology is not scalable, as in real world applications, the ontology is subject to constant change \cite{DBLP:conf/sigdial/HeckNLGLMG20}. In contrast, PPTOD directly generates the outputs, making it more adaptive and generalizable to new ontology labels in real world applications.


\begin{table}[tb]
    \small
	\centering  
	\renewcommand{\arraystretch}{1.2}
	\setlength{\tabcolsep}{6pt}
	\scalebox{0.78}{
	\begin{tabular}{ccccc}
		\hlinewd{0.75pt}
		\multirow{2}{*}{\textbf{Model}}&\multicolumn{4}{c}{Training Size (\%)}\\
		\cmidrule(lr){2-5}
		&1&5&10&20\\
		\hline
		SimpleTOD&7.91$\pm$1.07&16.14$\pm$1.48&22.37$\pm$1.17&31.22$\pm$2.32\\
		MinTL&9.25$\pm$2.33&21.28$\pm$1.94&30.32$\pm$2.14&35.96$\pm$1.25\\
		SOLOIST&13.21$\pm$1.97&26.53$\pm$1.62&32.42$\pm$1.13&38.68$\pm$0.98\\
		\hline
        $\textup{PPTOD}_{\textup{small}}$&27.85$\pm$0.77&39.07$\pm$0.85&42.36$\pm$0.29&45.98$\pm$0.38\\
        $\textup{PPTOD}_{\textup{base}}$&29.72$\pm$0.61&40.20$\pm$0.39&43.45$\pm$0.64&46.96$\pm$0.40\\
        $\textup{PPTOD}_{\textup{large}}$&\textbf{31.46$\pm$0.41}&\textbf{43.61$\pm$0.42}&\textbf{45.96$\pm$0.66}&\textbf{48.95$\pm$0.13}\\
		\hlinewd{0.75pt}
	\end{tabular}}
    \caption{Low-resource DST Evaluation: The means and standard deviations over five runs are reported.
    }
	\label{tb:few_shot_dst}
\end{table}

\subsubsection{Low-Resource Evaluation}
To investigate how well PPTOD performs with limited training samples on the downstream task, we evaluate it in a simulated low-resource setting. Specifically, we train the model on MultiWOZ 2.0 by varying the percentage of training data (i.e., 1\%, 5\%, 10\%, and 20\%). We compare PPTOD with three strong generation-based baselines, including SimpleTOD, MinTL, and SOLOIST, using the official code released by the authors.

Table \ref{tb:few_shot_dst} shows the experimental results. As seen, in all settings, PPTOD outperforms other baselines by a large margin. In the extreme scenario, with only 1\% of training data, PPTOD surpasses the strongest SOLOIST model by 18 points of accuracy. This demonstrates that our model is more generalizable and can be better applied to new tasks where the amount of training data is limited.

\begin{table*}[tb]
    \small
	\centering  
	\renewcommand{\arraystretch}{1.2}
	\setlength{\tabcolsep}{6pt}
	\scalebox{0.85}{
	\begin{tabular}{ccccccccc}
		\hlinewd{0.75pt}
		\multirow{2}{*}{\textbf{Model}}&\multirow{2}{*}{\textbf{Generation Mode}}&\multirow{2}{*}{\textbf{DB}}&\multicolumn{4}{c}{\textbf{End-to-End Dialogue Modelling}}&\multicolumn{2}{c}{\textbf{Inference Measurement}}\\
		\cmidrule(lr){4-7}
		\cmidrule(lr){8-9}
		&&&Inform$\uparrow$&Success$\uparrow$&BLEU$\uparrow$&Combined Score$\uparrow$&Latency (ms)$\downarrow$&Speedup$\uparrow$\\
		\hline
		SOLOIST&Cascaded&$\checkmark$&85.50&72.90&16.54&95.74&208.69&1.00$\times$\\
		MinTL&Cascaded&$\checkmark$&84.88&74.91&17.89&97.78&78.82&2.65$\times$\\
		\hline
		\multirow{4}{*}{T5-small}&\multirow{2}{*}{Cascaded}&$\times$&83.60&71.20&18.09&95.49&38.70&5.39$\times$\\
		&&$\checkmark$&84.10&73.70&18.03&96.93&39.78&5.25$\times$\\
		\cmidrule(lr){2-9}
		&\multirow{2}{*}{Plug-and-Play}&$\times$&84.70&72.80&\textbf{18.52}&97.27&\textbf{14.17}&\textbf{14.73}$\times$\\
		&&$\checkmark$&\textbf{85.10}&\textbf{75.10}&17.82&\textbf{97.92}&19.52&10.69$\times$\\
		\hlinewd{0.75pt}
	\end{tabular}}
    \caption{Comparison between plug-and-play and cascaded generation. $\uparrow$: higher is better and $\downarrow$: lower is better.}
	\label{tb:generation_mode_comparison}
\end{table*}


\begin{table}[tb]
    \small
	\centering  
	\renewcommand{\arraystretch}{1.2}
	\setlength{\tabcolsep}{6pt}
	\scalebox{0.9}{
	\begin{tabular}{cccc}
		\hlinewd{0.75pt}
		\multirow{2}{*}{\textbf{Model}}&\multicolumn{3}{c}{\# of Training Samples}\\
		\cmidrule(lr){2-4}
		&10&30&full\\
		\hline
		BERT-Fixed$\dagger$&67.55&80.07&87.19\\
		BERT-Tuned$\dagger$&83.42&90.03&93.66\\
		USE$\dagger$&84.23&89.74&92.81\\
		ConveRT$\dagger$&83.32&89.37&93.01\\
		USE+ConveRT$\dagger$&\textbf{85.19}&90.57&93.36\\
		SOLOIST$\ddagger$&78.73&89.28&93.80\\
		\hline
		$\textup{PPTOD}_{\textup{small}}$&78.87$\pm$0.36&87.88$\pm$0.26&93.27$\pm$0.39\\
		$\textup{PPTOD}_{\textup{base}}$&82.81$\pm$0.45&89.64$\pm$0.28&93.86$\pm$0.22\\
		$\textup{PPTOD}_{\textup{large}}$&84.12$\pm$0.23&\textbf{90.64$\pm$0.29}&\textbf{94.08$\pm$0.15}\\
		\hlinewd{0.75pt}
	\end{tabular}}
    \caption{Results on Banking77 dataset. $\dagger$ and $\ddagger$ are cited from \citet{DBLP:journals/corr/abs-2003-04807} and \citet{peng2021soloist}. 
    }
	\label{tb:intent_classification}
\end{table}

\subsection{Intent Classification}
The goal of intent classification, i.e. NLU, is to classify the user's intent based on the user's utterance. We conduct experiments on the benchmark Banking77 dataset \cite{DBLP:journals/corr/abs-2003-04807} that contains data with 77 different intents. Following previous studies \cite{DBLP:journals/corr/abs-2003-04807,peng2021soloist}, we test our model in both full training and low-resource settings. In the low-resource setting, we vary the number of training samples per intent from 10 to 30. The standard classification accuracy is reported for evaluation. 

We compare PPTOD with several strong baselines, including BERT-Fixed, BERT-Tuned, USE+ConveRT \cite{DBLP:journals/corr/abs-2003-04807}, USE \cite{DBLP:conf/acl/YangCAGLCAYTSSK20}, ConveRT \cite{DBLP:conf/emnlp/HendersonCMSWV20}, and SOLOIST \cite{peng2021soloist}. It is worth mentioning that all compared baselines are classification-based approach that uses a classifier with a softmax layer to make the prediction over the pre-defined intent set. In contrast, as described in section \cref{sec:multi-task-pretraining}, PPTOD solves the classification task as a generation problem by directly generating the text of intent label. Therefore, when adapting to a new classification task, PPTOD is more flexible and no extra model parameters are required.

\begin{table*}[t]
    \small
	\centering  
	\renewcommand{\arraystretch}{1.2}
	\setlength{\tabcolsep}{6pt}
	\scalebox{0.8}{
	\begin{tabular}{cccccccccccccc}
		\hlinewd{0.75pt}
		\multicolumn{4}{c}{\textbf{Pre-training Data Annotation}}&\multicolumn{6}{c}{\textbf{End-to-End Dialogue Modelling}}&\multicolumn{2}{c}{\textbf{Dialogue State Tracking}}&\multicolumn{2}{c}{\textbf{Intent Classification}}\\
		\cmidrule(lr){1-4}
        \cmidrule(lr){5-10}
        \cmidrule(lr){11-12}
        \cmidrule(lr){13-14}
		\multirow{2}{*}{NLU}&\multirow{2}{*}{DST}&\multirow{2}{*}{POL}&\multirow{2}{*}{NLG}&\multicolumn{3}{c}{1\% training}&\multicolumn{3}{c}{full training}&1\% training&full training&10 samples&full training\\
		\cmidrule(lr){5-7}
		\cmidrule(lr){8-10}
		\cmidrule(lr){11-11}
		\cmidrule(lr){12-12}
		\cmidrule(lr){13-13}
		\cmidrule(lr){14-14}
		&&&&Inform&Success&BLEU&Inform&Success&BLEU&Accuracy&Accuracy&Accuracy&Accuracy\\
		\hline
		$\times$&$\times$&$\times$&$\times$&53.28&36.08&11.65&83.10&72.40&18.17&17.44&50.55&75.12&92.91\\
		$\checkmark$&$\times$&$\times$&$\times$&58.58&40.48&11.02&85.20&73.50&16.96&18.47&50.71&78.21&\textbf{93.37}\\
		$\times$&$\checkmark$&$\times$&$\times$&66.10&46.40&11.26&86.30&74.90&18.52&\textbf{27.91}&51.48&75.97&93.03\\
		$\times$&$\times$&$\checkmark$&$\times$&60.60&48.20&11.88&84.40&74.60&18.55&19.32&50.82&75.37&92.95\\
		$\times$&$\times$&$\times$&$\checkmark$&59.38&40.78&12.34&83.60&74.70&\textbf{19.97}&17.82&50.58&75.61&92.97\\
		\hline
        $\checkmark$&$\checkmark$&$\checkmark$&$\checkmark$&\textbf{66.96}&\textbf{50.90}&\textbf{12.51}&\textbf{87.80}&\textbf{75.30}&19.89&27.85&\textbf{51.50}&\textbf{78.87}&93.27\\
		\hlinewd{0.75pt}
	\end{tabular}}
    \caption{Performance of models pre-trained on data with different annotations. In the low-resource setting of different tasks, the average scores over five runs are reported. The last row reports the results of $\textup{PPTOD}_{\textup{small}}$.
    }
	\label{tb:pretraining_investigation}
\end{table*}

In the experiments, we train PPTOD for five runs with different selection of training data and random seeds. The average scores and standard deviations are reported in Table \ref{tb:intent_classification}. We see that PPTOD is comparable with existing methods. On low-resource-30 and full training settings, $\textup{PPTOD}_{\textup{large}}$ achieves the best results. Our performance gains are even more remarkable given that PPTOD requires no extra parameters when solving the classification task.

\section{Further Analysis}
\label{sec:analysis}
In this section, we present further discussions and empirical analyses of the proposed model.

\subsection{Plug-and-Play vs Cascaded Generation}
First, we compare our plug-and-play generation with the cascaded generation that is adopted by most existing studies. 
To this end, we fine-tune a T5-small model (without dialogue multi-task pre-training) on MultiWOZ 2.0 by either using the plug-and-play or the cascaded formulation. Moreover, we also examine the effect of DB state on the model performance. Specifically, for the plug-and-play model, when utilizing DB state, it first predicts the dialogue state (DST) to retrieve the DB state from the pre-defined database. Then, based on the DB state and dialogue context, the output of POL and NLG are generated in parallel. When ignoring the DB state, the plug-and-play model generates DST, POL, and NLG results in a fully paralleled fashion.


For evaluation, we report the results on end-to-end dialogue modelling task. In addition, we report the average inference latency and relative speedup of each model.\footnote{The latency of each model is measured on a single Nvidia V100 GPU with a batch size of 1.} We compare our ablated models with two strong baselines, SOLOIST and MinTL.\footnote{We did not include TOP+NOD \cite{DBLP:journals/tacl/LiuYRB21} for comparison as the authors did not release their code.}

Table \ref{tb:generation_mode_comparison} presents the results. As seen, the plug-and-play models yield better results than their cascaded counterparts. One reason is that, for cascaded models, the previously generated results are explicitly used as model input for latter sub-tasks, which leads to error accumulation. Moreover, we see that using DB state generally improves the model performance for both plug-and-play and cascaded models as it provides the model with more grounding information. Furthermore, with DB state, our plug-and-play model achieves better overall score than MinTL with an around 4$\times$ speedup. This suggests that the plug-and-play formulation benefits the model both in terms of the generation accuracy as well as the inference latency.

\subsection{Multi-Task Pre-Training Investigation}
Next, we provide further analyses on the dialogue multi-task pre-training strategy. 
To quantify the importance of different pre-training data, we pre-train the T5-small model using data that is annotated for individual TOD-related task (i.e., NLU, DST, POL, and NLG). After pre-training, we then evaluate the models on three downstream TOD tasks using MultiWOZ 2.0 and Banking77 datasets. For end-to-end dialogue modelling and dialogue state tracking, we test the model in both 1\% and full training settings. For intent classification, we measure the accuracy of models trained with either 10 training samples per intent or full training samples. 

Table \ref{tb:pretraining_investigation} presents the results with the first row showing the performance of vanilla T5-small model. As seen, without any pre-training, the vanilla T5-small model performs poorly in the low-resource setting of all evaluated tasks. 
This suggests that the prior knowledge from pre-training is indispensable for the model to achieve strong performances in the low-resource scenarios. 

Moreover, we see that pre-training with data annotated for individual TOD-related task helps the model to attain better result in the corresponding downstream task. For example, pre-training with DST data notably improves the model performance in the downstream DST task both in low-resource and full-training settings. Similarly, pre-training with NLG data helps the model to get better BLEU score in the end-to-end dialogue modelling task. 

Lastly, we see that the $\textup{PPTOD}_{\textup{small}}$ model attains the best results on most of the evaluation metrics. This suggests that the pre-training data with different annotations are compatible with each other and the joint utilization of all pre-training data helps the model to achieve the best overall performance.

\subsection{Human Evaluation}
We also conduct a human evaluation with the help of graders proficient in English using an internal evaluation platform. For evaluation, we randomly selected 50 dialogue sessions from the test set of MultiWOZ 2.0 dataset. We compare the results generated by the $\textup{PPTOD}_{\textup{base}}$ model against the results from the SOLOIST model. All generated results, plus the reference, are evaluated by five graders on a 3-point Likert scale (0, 1, or 2) for each of the following features\footnote{More evaluation details are provided in the Appendix \ref{sec:human_evaluation_detail}.}:
\vspace{-1.5mm}
\begin{itemize}
    \item \textbf{Understanding}: Whether the system correctly understands the user's goal.
    \item \textbf{Truthfulness}: Whether the system's response is factually supported by the reference.\footnote{For this metric, we only evaluate the results of PPTOD and SOLOIST. By definition, the reference gets a score of 2.0.}
    \item \textbf{Coherency}: Whether the system's response is semantically coherent with the context.
    \item \textbf{Fluency}: Whether the system's response is grammatically fluent and easy to understand.
\end{itemize}

\begin{table}[tb]
    \small
	\centering  
	\renewcommand{\arraystretch}{1.2}
	\scalebox{0.78}{
	\begin{tabular}{ccccc}
		\hlinewd{0.75pt}
        &\textbf{Understanding}&\textbf{Truthfulness}&\textbf{Coherency}&\textbf{Fluency}\\
        \hline
        Agreement&0.641&0.598&0.668&0.806\\
        \hline
        Reference&1.92&2.00&1.93&1.98\\
        \hline
        SOLOIST&1.78&1.29&1.64&1.97\\
        PPTOD&\textbf{1.86}&\textbf{1.51}&\textbf{1.83}&\textbf{1.99}\\
		\hlinewd{0.75pt}
	\end{tabular}}
    \caption{Human Evaluation Results}
	\label{tb:human_evaluation}
\end{table}

Table \ref{tb:human_evaluation} lists the results, with the first row showing strong inter-annotator agreements as measured by Fleiss$\textprime$ kappa coefficient \cite{fleiss1971mns}. Comparing with SOLOIST, our model achieves better scores on all metrics. Moreover, on the truthfulness and coherency metrics, our model significantly outperforms SOLOIST as judged by Sign Test (p-value < 0.05), suggesting that PPTOD generates more factually correct and semantically coherent responses. Finally, we note that on the fluency metric, both systems perform comparably with the reference (p-value > 0.4). This shows that the fluency of such systems is largely guaranteed by the prior syntactic knowledge from pre-trained language models, which suggests that future research should focus more on the other aspects of dialog systems.

\section{Conclusion}
In this paper, we propose PPTOD, a unified model that supports both task-oriented dialogue understanding and response generation in a plug-and-play manner. In addition, we introduce a new dialogue multi-task pre-training strategy to further augment our model's ability in completing TOD-related tasks. Extensive experiments and analysis are conducted on three benchmark TOD tasks in both high-resource and low-resource settings. The automatic and human evaluations demonstrate that PPTOD outperforms the current SOTA systems in terms of various evaluation metrics.

\section*{Acknowledgments}
The authors would like to thank Anna Currey, David Vandyke, and Dingmin Wang for their insightful discussions and support. Many thanks to our anonymous reviewers and area chairs for their suggestions and comments.

\section*{Ethical Statement}
We honor and support the ACL code of Ethics. Task-oriented dialogue systems aim to  interact and assist the users to fulfill their goals. The interaction and assistance process do not involve any bias towards to the participants. All datasets used in this work are from previously published works, and in our view, do not have any attached privacy or ethical issues.

\bibliography{anthology,custom}
\bibliographystyle{acl_natbib}

\clearpage
\appendix

\section{Dataset Details}
\label{dataset_details}
We elaborate the details of the dialogue datasets contained in the pre-training dialogue corpora.

\begin{itemize}
    \item \textbf{MetaLWOZ} \cite{lee2019multi-domain} is designed for improving models' ability in generating natural language responses in unseen domains. It contains annotations for natural language generation (NLG) spanning over 47 domains.
    
    \item \textbf{SNIPS} \cite{DBLP:journals/corr/abs-1805-10190} is designed to help developing models capable of understanding users' intent (i.e., natural language understanding (NLU)). Its data consists of users' utterances gathered by crowdsourcing with over 20 intent labels across 9 domains. 
    
    \item \textbf{CLINC} \cite{larson-etal-2019-evaluation} is built for improving model's ability in detecting out-of-scope users' intents. It contains data with NLU annotations for 150 intents across 10 different domains. 
    
    \item \textbf{ATIS} \cite{ATIS} is used for building intent classification (NLU) model. It contains data with 22 user intents from the airline travel information domain.
    
    \item \textbf{KVRET} \cite{eric-etal-2017-key} is a in-car personal assistant dataset with dialogues from three domains: calendar scheduling, weather information retrieval, and point-of-interest navigation. It contains annotations for user belief state (DST) and system response (NLG).
    
    \item \textbf{WOZ} \cite{mrksic-etal-2017-neural} and \textbf{CamRest676} \cite{wen-etal-2017-network} are collected with Wizard-of-Oz procedure. They contains dialogues with DST and NLG annotations from the restaurant domain. 
    
    \item \textbf{MSR-E2E} \cite{li2018microsoft} contains dialogues from three domains, including  movie-ticket booking, restaurant reservation, and taxi booking. The data are annotated for three TOD-related tasks: DST, POL, and NLG.
    
    \item \textbf{Frames} \cite{el-asri-etal-2017-frames} contains dialogues from the trip booking domain. Its data are annotated for three TOD-related tasks, including DST, POL, and NLG.
    
    \item \textbf{TaskMaster} \cite{DBLP:conf/emnlp/ByrneKSNGDYDKC19} includes dialogues from six domains. Its data is collected with Wizard-of-Oz and self-dialogue approaches. The dataset is annotated with DST, POL, and NLG.
    
    \item \textbf{Schema-Guided} \cite{DBLP:conf/aaai/RastogiZSGK20} is used for the DSTC8 \cite{DBLP:journals/corr/abs-1911-06394} dialogue competition. It contains dialogues from 17 domains and it supports three TOD-related tasks, including DST, POL, and NLG.
\end{itemize}

\section{Low-Resource MultiWOZ Evaluation}
\label{sec:few-shot_mwoz_2_0_complete_results}
In Table \ref{tb:detailed_few_shot_multi_woz_2_0_experiment}, we show the results of our model on MultiWOZ 2.0 under different low-resource settings. To get more confident results, for each setting, we train our model for five runs with different selection of training data and different random seeds. The complete results along with the mean and standard deviations are presented in Table \ref{tb:detailed_few_shot_multi_woz_2_0_experiment}.

\begin{table*}[h]
    \small
	\centering  
	\renewcommand{\arraystretch}{1.2}
	\setlength{\tabcolsep}{6pt}
	\scalebox{0.75}{
	\begin{tabular}{ccccccccccccccccc}
		\hlinewd{0.75pt}
		\multirow{2}{*}{\textbf{Model}}&\multicolumn{4}{c}{1\% of training data}&\multicolumn{4}{c}{5\% of training data}&\multicolumn{4}{c}{10\% of training data}&\multicolumn{4}{c}{20\% of training data}\\
		\cmidrule(lr){2-5}
		\cmidrule(lr){6-9}
		\cmidrule(lr){10-13}
		\cmidrule(lr){14-17}
		&Inform&Succ.&BLEU&Comb.&Inform&Succ.&BLEU&Comb.&Inform&Succ.&BLEU&Comb.&Inform&Succ.&BLEU&Comb.\\
		\hline
		\multicolumn{17}{c}{$\textup{PPTOD}_{\textup{small}}$}\\
		\hline
		run-1&68.50&54.90&13.98&75.68&78.40&61.50&14.78&84.73&79.70&68.70&17.10&91.30&83.40&71.10&17.05&94.30\\
		run-2&64.70&50.20&12.19&69.64&75.20&61.30&15.85&84.10&87.00&67.30&13.89&91.04&82.80&68.90&17.03&92.88\\
		run-3&65.30&46.10&10.79&66.49&75.40&60.80&15.99&84.09&84.30&68.10&15.33&91.50&83.20&70.00&17.01&93.61\\
		run-4&64.80&51.00&12.43&70.33&77.20&59.70&15.75&84.20&84.50&71.90&14.51&92.71&82.40&69.40&17.93&93.83\\
		run-5&71.50&52.30&13.14&75.04&76.70&64.70&14.37&85.07&78.00&64.90&16.99&88.44&83.00&70.10&16.10&92.65\\
		\hline
		average&66.96&50.90&12.51&71.44&76.58&61.60&15.35&84.44&83.50&68.18&15.56&91.01&82.96&69.90&17.02&93.45\\
		std&2.67& 2.88&1.06&3.46&1.18&1.67&0.65&0.39&3.33&2.26&1.29&1.40&0.34&0.74&0.58&0.61\\

		\hline
		\multicolumn{17}{c}{$\textup{PPTOD}_{\textup{base}}$}\\
		\hline
		run-1&74.20&55.40&13.08&77.88&80.50&66.10&15.58&88.88&85.10&67.50&16.02&92.32&84.90&72.50&17.16&95.86\\
		run-2&71.20&51.10&13.32&74.47&81.50&63.10&14.32&86.62&84.60&69.00&15.06&91.86&84.00&72.50&16.46&94.71\\
		run-3&76.20&49.70&12.39&75.34&77.50&61.70&14.98&84.58&84.10&69.20&15.49&92.14&85.50&69.60&17.76&95.31\\
		run-4&75.80&52.40&13.21&77.30&79.70&62.30&15.13&86.10&84.40&68.30&15.17&91.52&84.20&70.70&16.88&94.33\\
		run-5&74.70&53.60&12.97&77.05&80.10&64.20&14.44&86.59&83.90&67.80&16.12&91.96&86.10&73.20&16.78&96.43\\
		\hline
		average&74.42&52.44&12.99&76.41&79.86&63.48&14.89&86.55&84.42&68.36&15.57&91.96&84.94&71.70&17.01&95.32\\
		std&1.76&1.97&0.32&1.29&1.32&1.55&0.46&1.38&0.42&0.66&0.43&0.27&0.79&1.34&0.44&0.75\\
		\hline
		\multicolumn{17}{c}{$\textup{PPTOD}_{\textup{large}}$}\\
		\hline
		run-1&64.40&51.90&11.30&69.45&75.20&59.80&14.01&81.51&79.30&64.60&14.82&86.77&82.10&69.70&14.68&90.58\\
		run-2&65.50&53.20&12.01&71.36&74.30&64.10&14.98&83.18&80.40&67.80&15.01&89.11&81.70&72.20&15.61&92.56\\
		run-3&66.20&50.80&11.94&70.49&76.90&62.30&14.01&83.61&81.30&69.20&16.23&91.48&80.90&70.80&14.33&90.18\\
		run-4&62.70&52.60&12.20&69.85&76.20&60.70&13.45&81.90&82.30&66.90&14.99&89.59&83.10&73.50&15.83&94.13\\
		run-5&63.10&51.20&11.73&68.88&73.40&62.80&14.42&82.52&79.90&65.20&15.21&87.76&80.90&74.70&15.21&93.01\\
		\hline
		average&64.38&51.94&11.84&70.01&75.20&61.94&14.17&82.54&80.64&66.74&15.25&88.94&81.74&72.18&15.13&92.09\\
		std&1.34&0.88&0.31&0.85&1.26&1.53&0.51&0.78&1.06&1.68&0.50&1.61&0.82&1.80&0.56&1.49\\
		\hlinewd{0.75pt}
	\end{tabular}}
    \caption{Low-Resource Experiments on MultiWOZ: The average and std rows show the mean and standard deviation of results from five different runs. The Succ. and Comb. denote Success and Combined Score, respectively.}
	\label{tb:detailed_few_shot_multi_woz_2_0_experiment}
\end{table*}

\section{Human Evaluation Guidelines}
\label{sec:human_evaluation_detail}
Please evaluate the system's response with respect to the following features: (1) Understanding; (2) Truthfulness; (3) Coherency; and (4) Fluency. In the following, we provide some guidelines regarding how to judge the quality of the system's response in terms of different features.

\subsection{Understanding}
This metric measures whether the system's response shows that the system is able to understand the goal and intent of the user. The definition of different scores are:
\begin{itemize}
    \item $2$: The system completely understands the user's goal and intent.
    \item $1$: The system partially understands the user's goal and intent.
    \item $0$: The system does not understand the user's goal and intent at all.
\end{itemize}

\subsection{Truthfulness}
This metric measures whether the system's response is factually supported by the reference response. The definition of different scores are:
\begin{itemize}
    \item $2$: The facts in the system's response are all supported by or can be inferred from the reference response.
    \item $1$: The facts in the system's response are partially supported by the reference response.
    \item $0$: The system's response is contradicted to the facts contained in the reference response.
\end{itemize}

\subsection{Coherency}
This metric measures whether the system's response is logically coherent with the dialogue context. The definition of different scores are:
\begin{itemize}
    \item $2$: The system's response is logically coherent with the dialogue context.
    \item $1$: The system's response contains minor information that is off the topic of the dialogue context.
    \item $0$: The system's response is  completely irrelevant to the dialogue context.
\end{itemize}

\subsection{Fluency}
The metrics measures the fluency of the system's response. The definition of different scores are:
\begin{itemize}
    \item $2$: The system's response is grammatically correct and easy to understand.
    \item $1$: The system's response contains minor errors but they do not affect your understanding.
    \item $0$: The system's response does not make sense and it is unreadable.
\end{itemize}

\section{Case Study}
Table \ref{tb:case_study} presents a generated dialogue example from the $\textup{PPTOD}_{\textup{base}}$ model. The user starts the conversation by asking for an expensive restaurant that serves Indian food for dinner. PPTOD finds 14 restaurants that satisfy the user's goal and asks the user for a preferred location. We can see that, when the user states no preference on the restaurant location, PPTPD correctly updates the dialogue state by adding the area information which is missed by the oracle information. Then the user switches the dialogue topic for booking a hotel. Through the dialogue trajectory, we see that PPTOD completes the dialogue by successfully providing the user the necessary information such as number of hotel choices (at turn 3) and the booking reference number (at turn 6). When finding the user's booking request cannot be fulfilled (at turn 5), the models asks the user for an alternative option. Moreover, this example also demonstrates that PPTOD is able to deal with some NLU challenges displayed in the conversations. For example, at turn 4, the user already provides the information about the Gonville Hotel. But only after the user describes the intention of booking the hotel at turn 5, the model updates the name of hotel in the dialogue state based on the co-referenced information from the previous turn. Interestingly, the hotel name is ignored by the oracle dialogue state but our model correctly detects it. The dialogue understanding ability of PPTOD can also be observed in turn 6, in which it updates the hotel stay in the belief state from 2 days to 1 day after the user provides the corresponding information.

\begin{table*}[tb]
    \small
	\centering  
	\renewcommand{\arraystretch}{1.3}
	\setlength{\tabcolsep}{6pt}
	\scalebox{0.9}{
	\begin{tabular}{cll}
		\hlinewd{0.75pt}
        \textbf{MUL0113}&&\textbf{Dialogue}\\
        \hlinewd{0.75pt}
        \multirow{6}{*}{Turn 1}&User Utterance&i am in need of an expensive indian restaurant for dinner .\\
        \cline{2-3}
        &Generated Belief State&[restaurant] \{food = indian, pricerange = expensive\}\\
        &Oracle Belief State&[restaurant] \{food = indian, pricerange = expensive\}\\
        \cline{2-3}
        &Delexicalized Response&\makecell[l]{there are [value_choice] [value_price] [value_food] restaurants in cambridge . what \\area would you like to dine in ?}\\
        &Lexicalized Response&\makecell[l]{there are 14 expensive indian restaurants in cambridge . what area would you like to\\ dine in ?}\\
        &Oracle Reference&there are 14 such restaurants . do you want to be in a specific part of town ?\\
        \hlinewd{0.75pt}
        
        \multirow{6}{*}{Turn 2}&User Utterance&i don't care what part of town it is in , i would like a phone number please .\\
        \cline{2-3}
        &Generated Belief State&[restaurant] \{food = indian, pricerange = expensive, \textcolor{red}{area = don't care}\}\\
        &Oracle Belief State&[restaurant] \{food = indian, pricerange = expensive\}\\
        \cline{2-3}
        &Delexicalized Response&\makecell[l]{[value_name] is a [value_food] restaurant in the [value_area] part of town . their phone\\ number is [value_phone] .}\\
        &Lexicalized Response&\makecell[l]{curry garden is a indian restaurant in the center part of town . their phone number is\\ 01223302330 .}\\
        &Oracle Reference& \makecell[l]{sure , i am sorry . i have curry garden their phone number is 01223302330 . would you \\like me to reserve a table for you ?}\\
        \hlinewd{0.75pt}
        
        \multirow{6}{*}{Turn 3}&User Utterance&not at this time . i am looking for a 3-star hotel , though . no guest house .\\
        \cline{2-3}
        &Generated Belief State&\makecell[l]{[hotel] \{stars = 3, type = hotel\}; [restaurant] \{food = indian, pricerange = expensive\}}\\
        &Oracle Belief State&[hotel] \{stars = 3, type = hotel\}; [restaurant] \{food = indian, pricerange = expensive\}\\
        \cline{2-3}
        &Delexicalized Response&\makecell[l]{i have [value_choice] [value_stars] star [value_type] in the [value_area] .  [value_name] \\is a [value_stars] star [value_type] . would you like to book a room ?}\\
        &Lexicalized Response&\makecell[l]{i have 2 3-star hotel in the center .  gonville hotel is a 3-star hotel . would you like to book\\ a room ?}\\
        &Oracle Reference&i have found the gonville hotel . it is a 3 star hotel . would you like me to book it for you ?\\
        \hlinewd{0.75pt}
        
        \multirow{6}{*}{Turn 4}&User Utterance&not yet . is the gonville hotel in the same price range as the curry garden ?\\
        \cline{2-3}
        &Generated Belief State&[hotel] \{stars = 3, type = hotel\}; [restaurant] \{food = indian, pricerange = expensive\}\\
        &Oracle Belief State&[hotel] \{stars = 3, type = hotel\}; [restaurant] \{food = indian, pricerange = expensive\}\\
        \cline{2-3}
        &Delexicalized Response&yes , it is in the [value_price] price range . would you like me to book it for you ?\\
        &Lexicalized Response&yes , it is in the expensive price range . would you like me to book it for you ?\\
        &Oracle Reference&yes , it is an expensive hotel .\\
        \hlinewd{0.75pt}
        
        \multirow{6}{*}{Turn 5}&User Utterance&can you book a room for 2 people for 2 nights starting on sunday ?\\
        \cline{2-3}
        &Generated Belief State&\makecell[l]{[hotel] \{stars = 3, type = hotel, \textcolor{red}{name = gonville hotel}, people = 2, day = sunday, stay = 2\};\\\ [restaurant] \{food = indian, pricerange = expensive\}}\\
        &Oracle Belief State&\makecell[l]{[hotel] \{stars = 3, type = hotel, people = 2, day = sunday, \textcolor{red}{stay = 2}\}; [restaurant] \{food = \\indian, pricerange = expensive\}}\\
        \cline{2-3}
        &Delexicalized Response&\makecell[l]{i am sorry , but the [value_name] is not available for that stay . would you like to try a \\shorter stay ?}\\
        &Lexicalized Response&\makecell[l]{i am sorry , but the gonville hotel is not available for that stay . would you like to try a \\shorter stay ?}\\
        &Oracle Reference&\makecell[l]{the booking was unsuccessful . would you like to try for another day or a shorter stay ?}\\
        \hlinewd{0.75pt}
        
        \multirow{6}{*}{Turn 6}&User Utterance&\makecell[l]{could you try it for 1 night ? if that works , please book it and get the reference number .}\\
        \cline{2-3}
        &Generated Belief State&\makecell[l]{[hotel] \{stars = 3, type = hotel, \textcolor{red}{name = gonville hotel}, people = 2, day = sunday, \textcolor{red}{stay = 1}\};\\\ [restaurant] \{food = indian, pricerange = expensive\}}\\
        &Oracle Belief State&\makecell[l]{[hotel] \{stars = 3, type = hotel, people = 2, day = sunday, stay = 1\}; [restaurant] \{food = \\indian, pricerange = expensive\}}\\
        \cline{2-3}
        &Delexicalized Response&\makecell[l]{i was able to successfully book your room . your reference number is [value_reference] . \\is there anything else i can help you with ?}\\
        &Lexicalized Response&\makecell[l]{i was able to successfully book your room . your reference number is 7rzme01z . is there\\ anything else i can help you with ?}\\
        &Oracle Reference&\makecell[l]{you are all set for 1 night ( sunday ) , reference number is 6wvv053q . would you like to \\try another location for monday night ?}\\
        \hlinewd{0.75pt}
        
        \multirow{6}{*}{Turn 7}&User Utterance&no . that is all i need , thank you .\\
        \cline{2-3}
        &Generated Belief State&\makecell[l]{[hotel] \{stars = 3, type = hotel, name = gonville hotel, people = 2, day = sunday, stay = 1\};\\\ [restaurant] \{food = indian, pricerange = expensive\}}\\
        &Oracle Belief State&\makecell[l]{[hotel] \{stars = 3, type = hotel, people = 2, day = sunday, stay = 1\}; [restaurant] \{food = \\indian, pricerange = expensive\}}\\
        \cline{2-3}
        &Delexicalized Response&you are welcome . have a great day !\\
        &Lexicalized Response&you are welcome . have a great day !\\
        &Oracle Reference&thank you , and goodbye .\\
        \hlinewd{0.75pt}
	\end{tabular}}
    \caption{An generated dialogue example from the $\textup{PPTOD}_{\textup{base}}$ model. (best viewed in color)}
	\label{tb:case_study}
\end{table*}

\end{document}